\documentclass[10pt,twocolumn,letterpaper]{article}

\usepackage{3dv}
\usepackage{times}
\usepackage{epsfig}
\usepackage{graphicx}
\usepackage{amsmath}
\usepackage{amssymb}

\usepackage{subfigure}
\usepackage{multirow}
\usepackage{enumitem}
\usepackage{url}

\usepackage{xcolor}

\usepackage[pagebackref=true,breaklinks=true,colorlinks,bookmarks=false]{hyperref}

\definecolor{somegray}{rgb}{0.5, 0.5, 0.5}
\newcommand{\darkgrayed}[1]{\textcolor{somegray}{#1}}
\makeatletter
\newcommand*\titleheader[1]{\gdef\@titleheader{#1}}
\AtBeginDocument{%
  \let\st@red@title\@title
  \def\@title{%
    \vskip-3em
    \bgroup\normalfont\large\centering\@titleheader\par\egroup
    \vskip1.5em\st@red@title}
}
\makeatother

\threedvfinalcopy % *** Uncomment this line for the final submission

\def\threedvPaperID{419} % *** Enter the 3DV Paper ID here

\ifthreedvfinal\fi
\titleheader{\darkgrayed{This paper has been accepted for publication at the\\
IEEE International Conference on 3D Vision (3DV), 2021.
\copyright IEEE}}

\title{Object SLAM-Based Active Mapping and Robotic Grasping}
\begin{document}
%\and
\author{Yanmin Wu$^{1}$,\quad Yunzhou Zhang$^{1}$\thanks{\,The corresponding author of this paper. \newline \text{\quad\quad Email: {\tt\small zhangyunzhou@mail.neu.edu.cn}}},\quad Delong Zhu$^{2}$,\quad Xin Chen$^{1}$,\quad Sonya Coleman$^{3}$, \\ Wenkai Sun$^{1}$,\quad Xinggang Hu$^{1}$,\quad Zhiqiang Deng$^{1}$
	\vspace{0.2cm}\\
	$^1$Northeastern University \quad$^2$The
Chinese University of Hong Kong \quad$^3$Ulster University
	\vspace{-0.2cm}\\
	%\thanks{\textsuperscript{*}The corresponding author of this paper.}
}

\maketitle
\thispagestyle{empty}

%%%%%%%%% ABSTRACT
\begin{abstract}
   This paper presents the first active object mapping framework for complex robotic manipulation and autonomous perception tasks. The framework is built on an object SLAM system integrated with a simultaneous multi-object pose estimation process that is optimized for robotic grasping. Aiming to reduce the observation uncertainty on target objects and increase their pose estimation accuracy, we also design an object-driven exploration strategy to guide the object mapping process, enabling autonomous mapping and high-level perception. Combining the mapping module and the exploration strategy, an accurate object map that is compatible with robotic grasping can be generated. Additionally, quantitative evaluations also indicate that the proposed framework has a very high mapping accuracy. Experiments with manipulation (including object grasping and placement) and augmented reality significantly demonstrate the effectiveness and advantages of our proposed framework. 
\end{abstract}

%%%%%%%%% BODY TEXT
\vspace{-4mm}
\section{Introduction}
\vspace{-2mm}
%\vspace*{-0.5\baselineskip}

Active mapping refers to altering the robot's state by some specific criteria to autonomously perceiving the environment \cite{zhu2018deep}, which has wide applications in mobile robots, e.g., unfamiliar environment exploration \cite{chen2020active, hawkeye-zhu}, object searching \cite{zeng2020semantic}, and navigation \cite{ramakrishnan2020occupancy, cimurs2021goal}. In recent years, with the growing requirements on robotic manipulation in unknown settings, active mapping also receives much attention in the manipulation community \cite{wada2020morefusion, sucar2020nodeslam, labbe2020cosypose, almeida2019detection}.

However, the output of the conventional mapping framework, e.g., KinectFusion \cite{izadi2011kinectfusion}, ORB-SLAM2 \cite{mur2017orb}, cannot be used directly for manipulation tasks like object grasping, since the generated map does not contain object pose information. Therefore, additional processing is required to register the object pose into the map. This is often performed by aligning the CAD model with the object point cloud \cite{wada2020morefusion, sucar2020nodeslam, labbe2020cosypose} or directly predicting the object pose using deep neural networks \cite{tremblay2018deep, sucar2020nodeslam,wen2020se}, typically limited to specific scenarios and objects. Unsolved is the challenge of estimating the pose of novel objects, which is critical in the context of universal robot operation. Another alternative solution \cite{morrison2018closing,fang2020graspnet,kumra2020antipodal} focuses on grip detection, achieving a high grasp accuracy for novel objects and cluttered settings. However, they only concern the grasping, which is insufficient for intelligent robot tasks such as object delivery requested by the user and object placement, due to the absence of object-level perception. Thus, the observation of this paper is that \textbf{intelligent robot manipulation requires the ability of global perception, which encompasses both general objects and their surrounding}. Therefore, the generic object-oriented SLAM \cite{nicholson2018quadricslam,yang2019cubeslam,wu2020eao} is a workable option that estimates the object pose and builds the global map simultaneously. Additionally, the intelligent and autonomous robot should actively perceive its environment without human intervention. The goal of the majority of studies \cite{kriegel2015efficient,arruda2016active,monica2018contour} is to map the environment completely, and the active strategy is determined by the map's completeness, which is illogical given that many areas are invalid for object pose estimation. Therefore, another discovery made in this study is that \textbf{the active mapping strategy used in the object manipulation application should be object-driven}.

To address these problems, we propose an active mapping framework for robotic grasping and object manipulation. The framework approximates unknown objects using cylinders and cubes and combines object pose estimation into a joint optimization process, achieving simultaneous object and camera pose (the camera is fixed with the robot arm manipulator) estimation to realize global perception. In addition, we suggest an object-drive exploration strategy to actively perceive the environment to reduce the uncertainty introduced by incomplete observation and noise associated with object pose estimation.

The contributions are summarized as follows:
%[topsep=5pt,itemsep=-1ex,partopsep=1ex,parsep=1ex]
\begin{itemize}[topsep=5pt,itemsep=-1ex,partopsep=1ex,parsep=1ex]
	\item We extend the object pose estimation algorithm based on our previous work \cite{wu2020eao}, making it more robust and accurate, suitable for robotic grasping.	
	\item We present an object-driven exploration strategy that takes into consideration the completeness of object observations and the uncertainty associated with pose estimate, which significantly improves the accuracy of the actively-generated object map.
	\item To the best of our knowledge, this is the first time an object SLAM  is combined with an object-driven exploration strategy, achieving extremely accurate object mapping, complex robotic grasping, and object placement tasks. Additional experiments and videos are available on our project page \url{https://yanmin-wu.github.io/project/active-mapping/}.
\end{itemize}

\section{Related Work}
\subsection{Active Perception and Mapping}  
%Active perception can be broadly categorized into geometry-based methods and learning-based methods. 
Active perception is the process of actively adjusting sensor states by analyzing existing data to obtain more valuable information. Zhang \textit{et al.} \cite{zhang2019beyond} leverage Fisher information to predict the best sensing location for reducing localization uncertainty. Zeng \textit{et al.} \cite{zeng2020semantic} leverage prior knowledge between objects to establish a semantic link graph for active object searching. Zhu \textit{et al.} \cite{zhu2018deep, li2019Learning} propose to learn global topological knowledge of indoor environments using deep reinforcement learning to accelerate the efficiency of robotic exploration.
Active mapping is a specific task of active perception. Charrow \textit{et al.} \cite{charrow2015information} utilize the quadratic mutual information to guide 3D dense map building. Wang \textit{et al.} \cite{wang2019srm} also leverage the mutual information to perform Next-Best View (NBV) selection on a sparse road map, which then serves as the semantic landmark to help accelerate the mapping process. Kriegel \textit{et al.} \cite{kriegel2015efficient} propose a surface reconstruction method for single unknown objects. In addition to the information gain, the authors also integrate the measurement of reconstruction quality to the objective function, achieving very high accuracy and completeness.  

The data-driven exploration strategy proposed in this work is implemented based on the information entropy but additionally integrates the requirement for accurate object pose estimation. Another major difference with other methods is that the output of our proposed method is an object map that is compatible with complex grasping tasks.

\subsection{Object Map for Grasping}
An object map is an indispensable module for complex grasping tasks, e.g., object placement and arrangement. Wada \textit{et al.} \cite{wada2020morefusion} propose to reconstruct objects by incremental object-level voxel mapping. The object pose is initialized by voxel points, but there is a large error in scale. The ICP algorithm then is used to align the initialized object with the CAD model to further optimize the pose, the performance of which highly depends on the registration accuracy of the CAD model. Sucar \textit{et al.} \cite{sucar2020nodeslam} propose a multi-category object descriptor and a probabilistic rendering model to infer object poses and shapes. The target object is regarded as a landmark in SLAM and is involved in the joint optimization to help generate an accurate object map. The major deficiency of this method is that the model requires a tedious category-level training process for each object. Labb{\'e} \textit{et al.} \cite{labbe2020cosypose} propose a single-view 6-DoF object pose estimation method and utilize the object-level bundle adjustment in the SLAM framework to optimize the object map. However, their proposed method is only targeted for known objects. Almeida \textit{et al.} \cite{almeida2019detection} leverage the SLAM framework to densely map unknown objects for accurate grasping point detection, but the object pose is not estimated.

We also use the SLAM framework to generate the object map used for complicated grasping tasks. However, unlike previous studies, we focus on the pose estimation of unknown objects in the environment and leverage active mapping techniques to generate a more accurate and complete object map. To the best of our knowledge, this is the first work that applies active mapping to object pose estimation and unknown object grasping.

\vspace{-2mm}
\section{System Overview}
\vspace{-2mm}

\begin{figure}[t]
	\centering
	\includegraphics[width=0.47\textwidth]{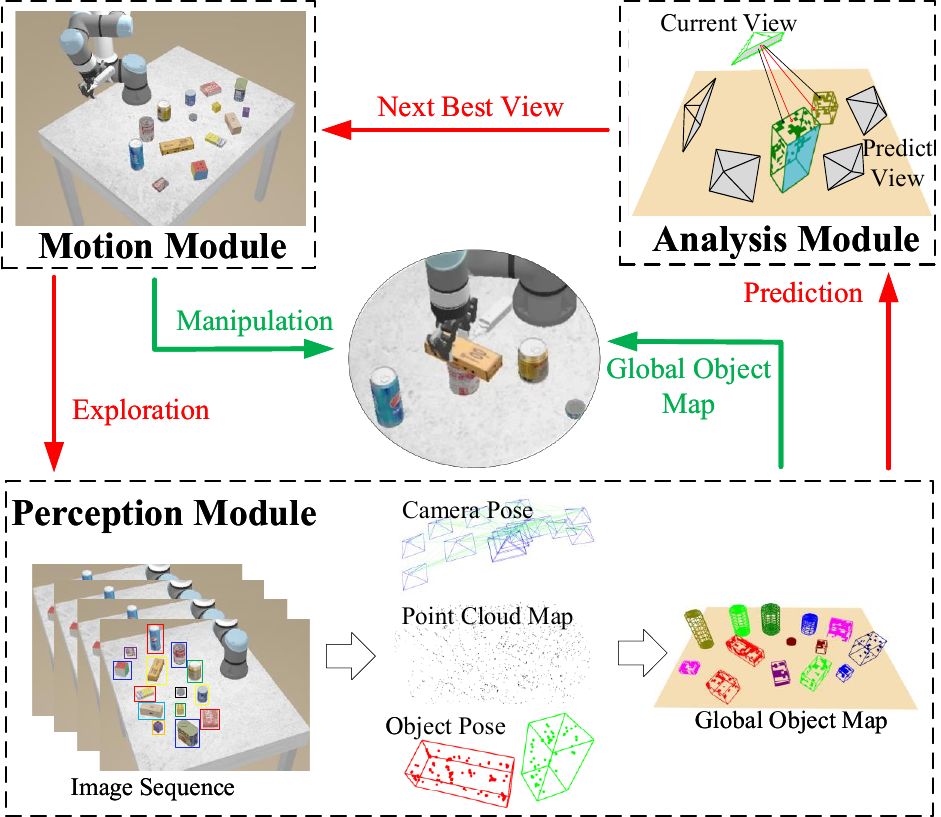}
	\caption{The proposed system framework. Motion module: Control the robot movement. Perception module: Create the global object map. Analysis module: Select the next best observation view by analyzing the object map.}
	\label{System_Framework}
\vspace{-5mm}
\end{figure}

An overview of the proposed system is presented in Fig. \ref{System_Framework}, which forms a closed-loop for incrementally exploring and mapping the desk environment. Specifically, the motion module is utilized to control the robot equipped with a camera as it executes observation commands. The perception module is used to estimate the camera pose and build the point cloud map, and most importantly, extract objects from the map and estimate their poses. The output of this module is an object map with the 9-DoF (position, orientation, and scale) object pose registered, which assists the robot in performing tasks such as motion planning and object manipulation. The analysis module quantifies object uncertainty and predicts the information gain of different camera views. The view with the largest information gain is selected as the NBV and passed to the motion module to enable active exploration. This work intends to incrementally develop a global object map with minimum effort and maximum accuracy for robotic grasping. Moreover, the produced object map enables the global perception for high-level robot manipulation tasks. Section \ref{section: Object Map} will illustrates the perception module, while sections \ref{section: Observation} and \ref{section: Exploration} will demonstrate the analysis and motion modules. The manipulation tasks will be provided in section \ref{section: Object Grasping} and \ref{section: Object Placement}.

\section{Object Driven Active Mapping}
\vspace{-1mm}
\subsection{Object Map Building}
\label{section: Object Map}

Point cloud map permits camera localization, while the multiple object pose provides object-level perception. The object map refers to register multi-objects to the point cloud map to achieve global geometric and semantic perception. In which, the object pose is critical for successful grasping, and the global perception enables robot high-level tasks such as planning and decision-making. The main components of the object map are object-level data association, object pose estimation, and optimization. 

\begin{figure}[t]
	\centering
	%\captionsetup{belowskip=-10pt}
	\includegraphics[scale=0.28]{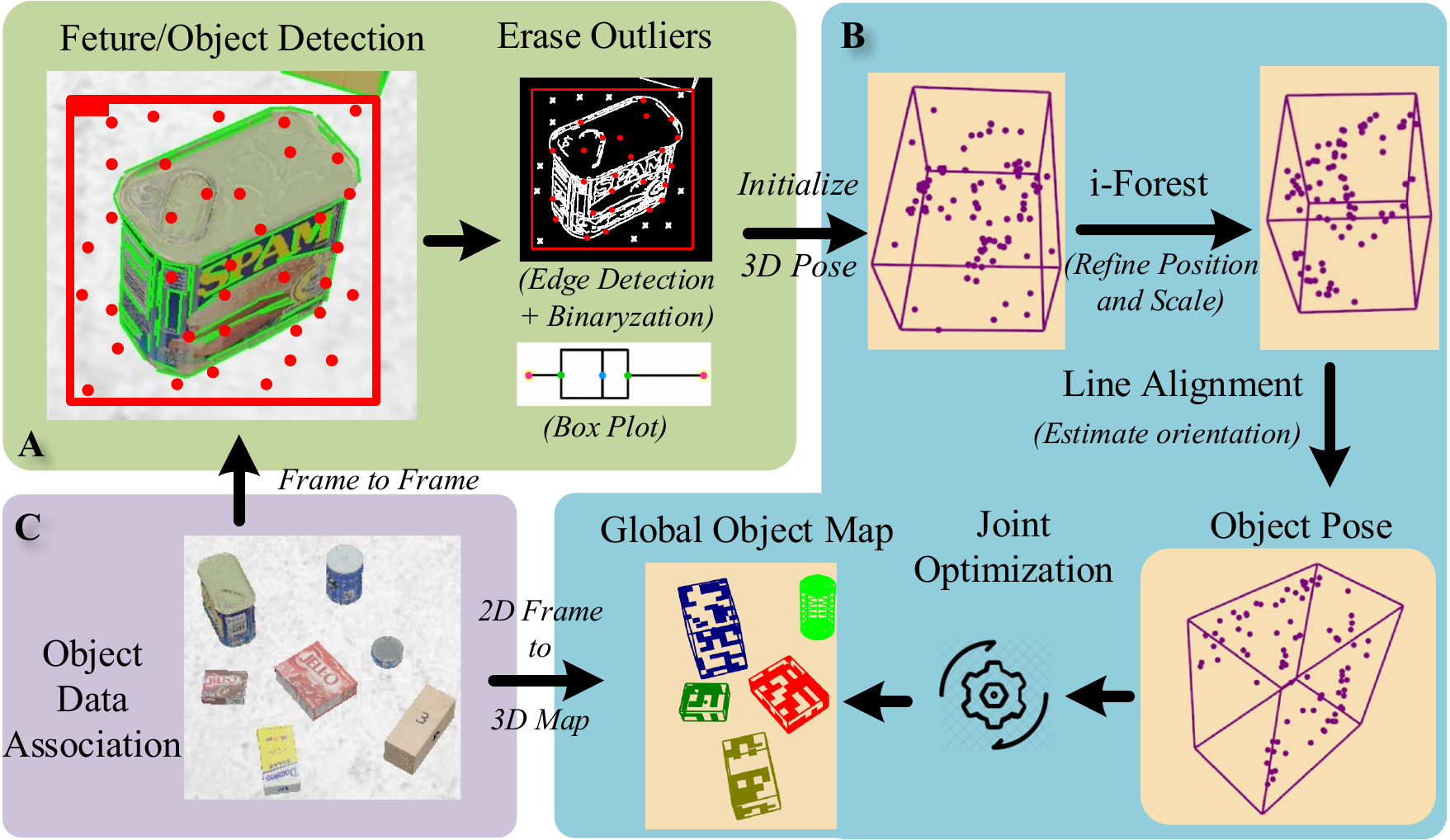}
	\caption{The framework of object pose estimation. A: associate point clouds with the object and remove outliers. B: estimate the object pose and create the object map. C: object-level data association in multi-frame. For more details, refer to \cite{wu2020eao}.}
	\label{Pose Estimation}
	\vspace{-1mm}
\end{figure}

As demonstrated in Fig. \ref{Pose Estimation}, the point-line features and objects are first detected on each image frame. After removing the background points for each object, The remaining points are gathered in the corresponding object-point cloud in the global map. The association of objects in different images is discovered following a strong object association technique in \cite{wu2020eao}, and the object pose is initialized by fitting a tight 3D cube to the object point cloud using the i-Forest \cite{liu2012isolation} and line alignment. Then, a joint optimization of the camera and object pose is performed to further improve the pose estimation accuracy. The above-mentioned technical details can be found in our previous work \cite{wu2020eao}. In this work, we focus on proposing a new optimization algorithm that can achieve highly accurate object pose estimation.

\begin{figure}[bt]
	\centering
%	\captionsetup{belowskip=-10pt}
	\includegraphics[scale=0.27]{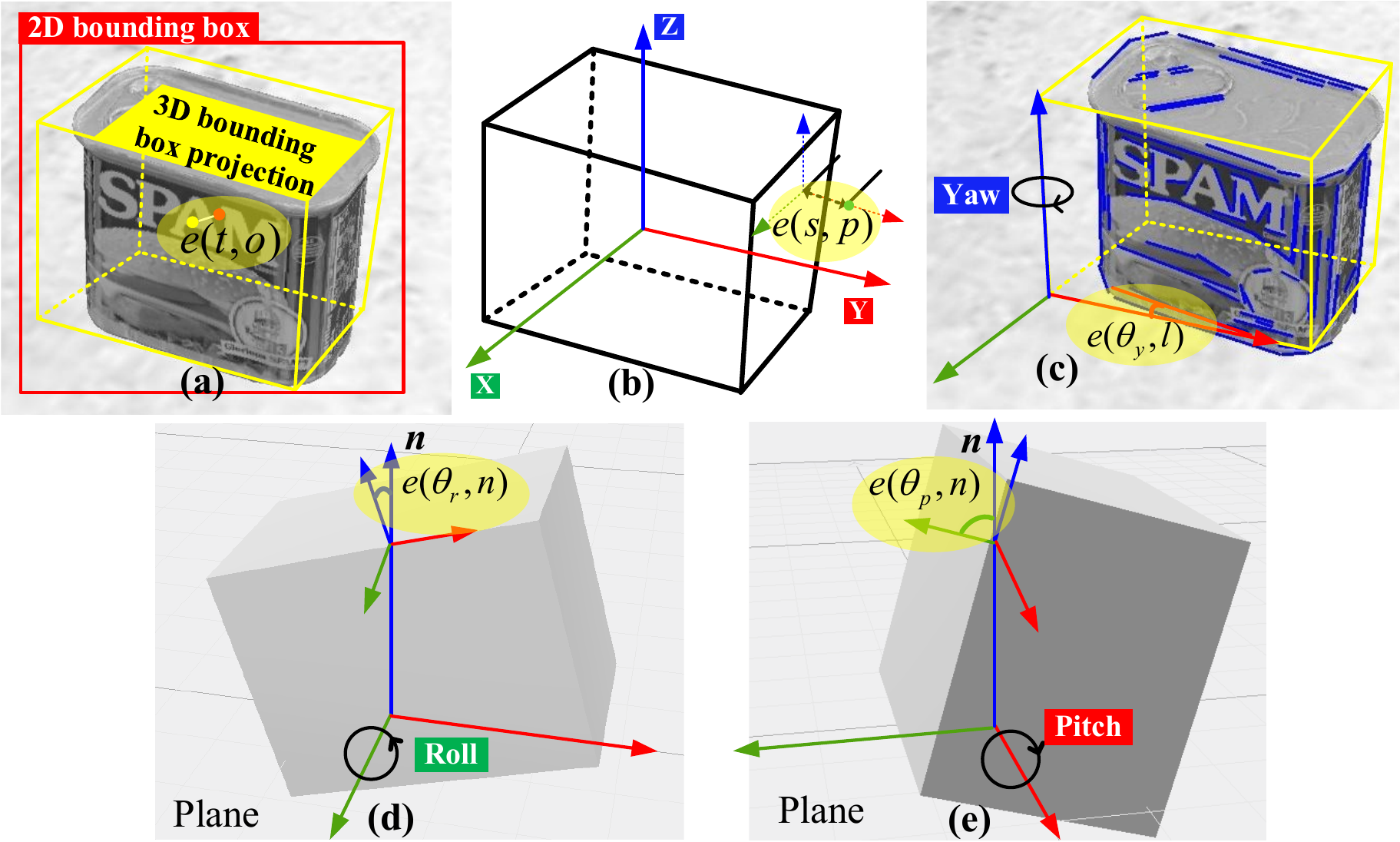}
	\caption{Error definition of object pose optimization.}
	\label{bject_pose_optimization}
	\vspace{-6mm}
\end{figure}

The object pose is parameterized with $O=\{\boldsymbol{t}, \boldsymbol{\theta}, \boldsymbol{s}\}$, which represents the translation, orientation, and scale of the object, respectively. The error function used to optimize the object pose is formulated as follows, %LM algorithm was adopted to solve the nonlinear optimization problem:
\vspace*{-0.5\baselineskip}
\begin{equation}
\small
\arg \min_{O} \sum\left(e(\boldsymbol{t}, o)+e(\boldsymbol{s}, \boldsymbol{p})+e\left(\theta_{y}, l\right)+e\left(\left(\theta_{r}, \theta_{p}\right), \boldsymbol{n}\right)\right),
\label{opt}
\end{equation}
which is then solved by nonlinear optimizers, e.g., the Levenberg-Marquardt method. 

The first item in Eq. \eqref{opt} is the position error. As shown in Fig. \ref{bject_pose_optimization}(a), the 3D center $\boldsymbol{t}$ of the object is projected onto the image plane, and the distance between the projected point and the center of 2D bounding box $o$ is taken as the first error,
\vspace*{-0.5\baselineskip}
\begin{equation}
e(\boldsymbol{t}, o)=||\pi\left(T_{c}^{-1} \boldsymbol{t}\right)-o||_2,
\end{equation}
%\vspace*{-0.5\baselineskip}
where $\pi$ is the projection matrix, $T_{c}$ is the camera pose.

The second item is the scale error. As shown in Fig. \ref{bject_pose_optimization}(b), similar to the idea of CubeSLAM \cite{yang2019cubeslam}, the 3D object points $\boldsymbol{p}$ should be constrained inside the 3D cube, hence the distance from the point $\boldsymbol{p}$ to the corresponding cube surface is taken as the error,
\vspace*{-0.5\baselineskip}
\begin{equation}
e(\boldsymbol{s}, \boldsymbol{p})=\max \left(\left|T_{o}^{-1} \boldsymbol{p}\right|-\boldsymbol{s}, 0\right),
\end{equation}
where $T_{o}$ is the object pose.

The third item is the error between the yaw $\theta_{y}$ of the object and the direction $\theta_{l}$ of line $l$ in the image, shown in Fig. \ref{bject_pose_optimization}(c). The direction of the object should be parallel with the detected lines, and thus the angle should be minimized,
\vspace*{-0.5\baselineskip}
\begin{equation}
\label{yaw1}
e\left(\theta_{y}, l\right)=||\theta_{y}-\theta_{l}||_2.
\end{equation}

The fourth item is the roll and pitch errors (see Fig. \ref{bject_pose_optimization}(d) and (e)), which is omitted in CubeSLAM since the initial camera pose is assumed to be parallel to the ground plane, which in the grasping setting cannot be guaranteed. For the grasping task, objects are placed flat on the desktop, the normal vector $\boldsymbol{n}$ of which can be obtained by a RANSAC-based plane fitting algorithm \cite{fischler1981random}. The pitch of the object should be perpendicular to $\boldsymbol{n}$, while the roll should be parallel with $\boldsymbol{n}$,
\vspace*{-0.5\baselineskip}
\begin{equation}
\label{yaw2}
\begin{array}{c}
e\left(\theta_{r}, \boldsymbol{n}\right)=||\theta_{n}-\theta_{r}||_2, \\
e\left(\theta_{p}, \boldsymbol{n}\right)=||\left(\theta_{n}-\theta_{p}\right)-90^{\circ}||_2.
\end{array}
\end{equation}

By replacing the joint optimization module in Fig. \ref{Pose Estimation} with the new optimization algorithm, the accuracy of object pose estimation can be further improved, as well as the perception ability of the object map.

\subsection{Observation Completeness Measurement}
\label{section: Observation}

\label{OIM}
\begin{figure}[t]
	\centering
%		\captionsetup{belowskip=-10pt}
	\includegraphics[scale=0.26]{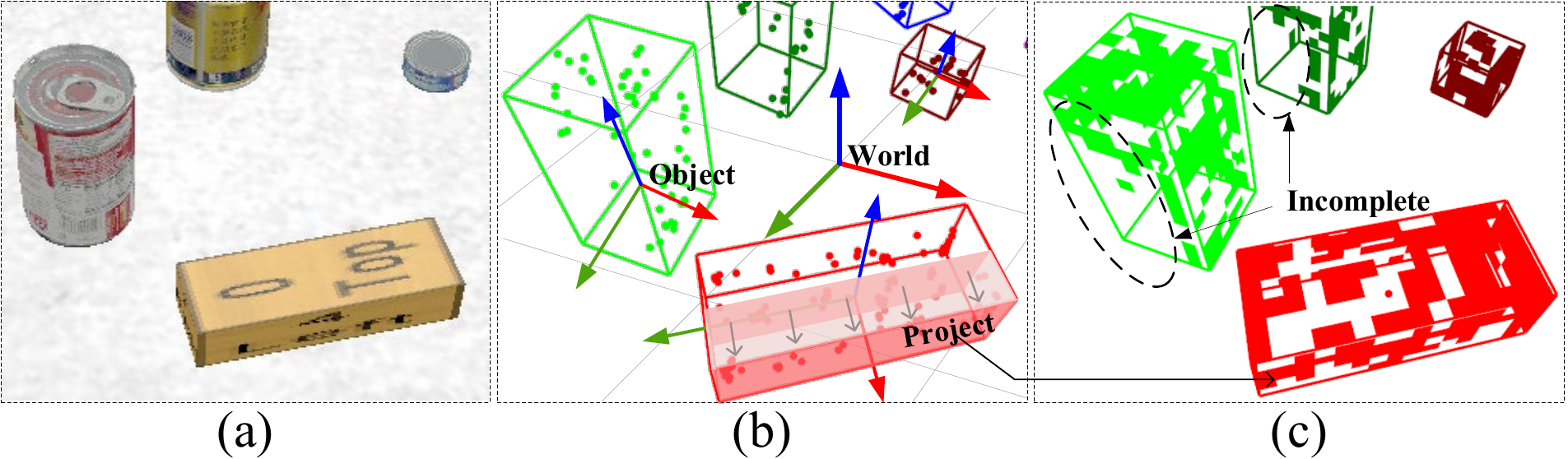}
	\caption{Observation completeness measurement. {(a): Raw image. (b): Objects with point cloud. (c): Objects with surface grids.}}
	\label{Object Incompleteness}
	\vspace{-4mm}
\end{figure}

In this work, we focus on the robot active map building. We regard the map's incompleteness as a motivating factor for active exploration. Existing studies usually take the entire environment as the exploration target \cite{charrow2015information, kahn2015active} or focus on reconstructing a single object \cite{kriegel2015efficient, arruda2016active}, neither of which is ideal for building the object map required by robotic grasping. The reasons are as follows: 1) The insignificant regions in the environment will interfere with the decisions made for exploration and misguide the robot into the non-object area; 2) it will significantly increase the computational cost and thus reduce the efficiency of the whole system. We propose an object-driven active exploration strategy for building the object map incrementally. The strategy is designed based on observation completeness of the object, which is defined as follows.

As illustrated in Fig. \ref{Object Incompleteness}, the point cloud of an object is translated from the world frame to the object frame and then projected onto the five surfaces of the estimated 3D cube (see Fig. \ref{Object Incompleteness}(b)). Here, only half of the points are projected to a single surface, and the bottom face is not considered. Each of the five surfaces is discretized into a surface occupancy grid map \cite{elfes1989using,tompkins2020online} with $m*m$ resolution ($m=1\rm cm$ in our implementation) (see Fig. \ref{Object Incompleteness}(c)). Each grid cell can be in one of three states:
\begin{itemize}[topsep=5pt,itemsep=-1ex,partopsep=1ex,parsep=1ex]
	\item \textbf{unknown}: the grid is not observed by the camera;
	\item \textbf{occupied}: the grid is occupied by the point clouds;
	\item \textbf{free}: the grid can be seen by the camera but is not occupied by the point clouds.
\end{itemize}

We use information entropy \cite{shannon1948mathematical} to determine the completeness of observations based on the occupancy grid map, as information entropy has the property of symptomatizing uncertainty.
The entropy of each grid cell is defined by a binary entropy function:
\vspace*{-0.5\baselineskip}
\begin{equation}
\label{eq:grid}
H_{gr i d}(p)=-p \log (p)-(1-p) \log (1-p),
\end{equation}
where $p$ is the probability of a grid cell being occupied and its initial value before exploration is set to 0.5 (at which value the entropy is maximal). The total entropy is then defined as
\vspace*{-0.5\baselineskip}
\begin{equation}
\label{eq:obj}
H_{o b j}=\sum_{o\in\mathbb{O}} H_{o}+\sum_{f \in \mathbb{F}} H_{f}+\sum_{u \in \mathbb{U}} H_{u},
\end{equation}
and the normalized total entropy is
\vspace*{-0.5\baselineskip}
\begin{equation}
\label{eq:ba_grid}
\bar{H}_{obj}=H_{o b j} /(|\mathbb{O}|+|\mathbb{F}|+|\mathbb{U}|),
\end{equation}
where $\mathbb{O}, \mathbb{F}, \mathbb{U}$ are sets of the occupied, free, and unknown grid cells, respectively. $|\mathbb{X}|$ represents the size of $\mathbb{X}$. As objects continue to be explored, the number of unknown grid cells ($p=0.5$) gradually decreases, the probability of an occupied grid increases increasingly to 1, and the probability of a free grid reduces gradually to 0, resulting in a decreasing value for $\bar{H}_{g ri d}$. The lower the $\bar{H}_{g ri d}$ is, the higher the observation completeness is. The exploration objective is to minimize $\bar{H}_{g ri d}$.

\begin{figure}[t]
	\centering
%	\captionsetup{belowskip=-10pt}
	\includegraphics[scale=0.25]{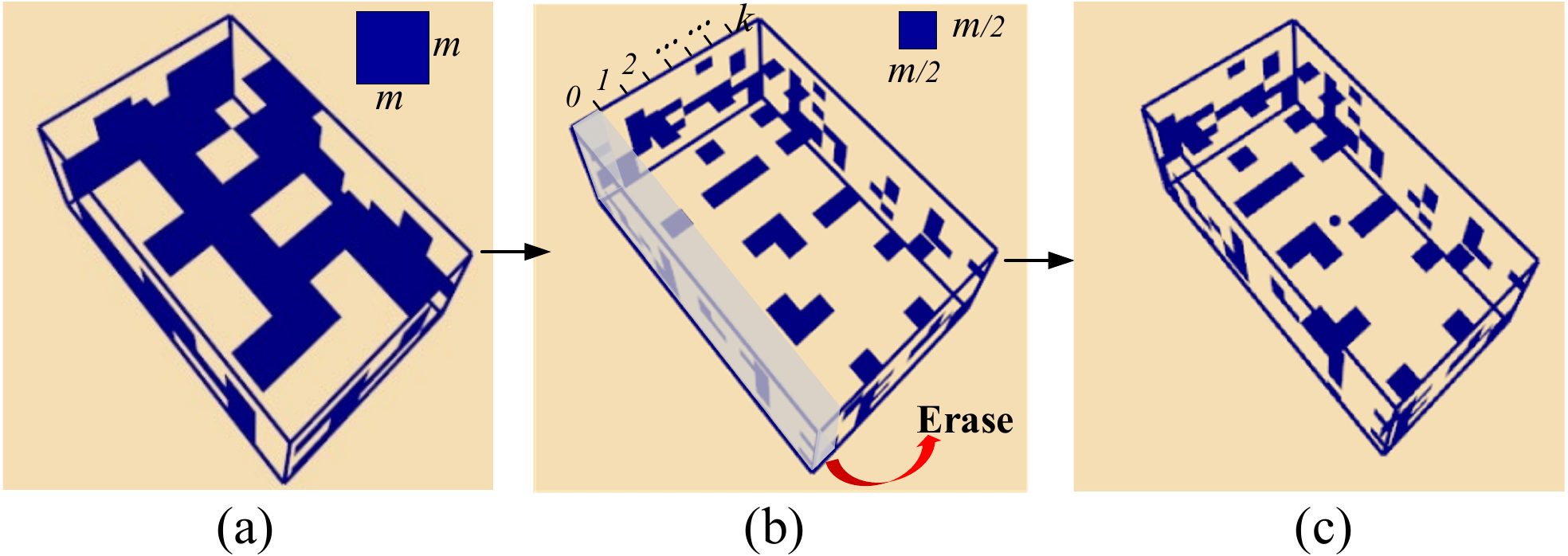}
	\caption{Pose refinement based on surface grid sparsity.}
	\label{Pose Refine}
	\vspace{-4mm}
\end{figure}

Additionally, based on the definition of the surface grid, an outlier removal method is also implemented, which is added after the i-Forest algorithm in Fig. \ref{Pose Estimation}. The idea is illustrated in Fig. \ref{Pose Refine}. To begin, reduce the resolution of the surface grid from $m$ to $m/2$ to obtain a more accurate pose (see Fig. \ref{Pose Refine}(a) to Fig. \ref{Pose Refine}(b)). Then, the 3D cube is divided into slices in one of the axes, and the number of occupied grid cells in each slice $(n_0, n_1, \cdots, n_K)$ is counted. Edge slices, e.g., {slice $0$ and $K$, that satisfy $n_i < 1/3 n_{i+1}$ ($0 < i < K/2$) or $n_i < 1/3 n_{i-1}$ ($K/2 < i < k$)} are ignored (see Fig. \ref{Pose Refine}(b) to Fig. \ref{Pose Refine}(c)). Such processing is conducted on all three axes. While this method is comparable to the i-Forest algorithm, it outperforms it in removing clustered outliers, which are difficult to spot using the i-Forest method.

\subsection{Object-Driven Exploration}
\label{section: Exploration}

\begin{figure}[t]
	\centering
	\subfigure[Different definitions of information gain in exploration. Left: target scene. Middle: uncertainty region (green) of the standard method. Right: uncertainty area (green) of our object-driven strategy.]{\includegraphics[scale=0.59]{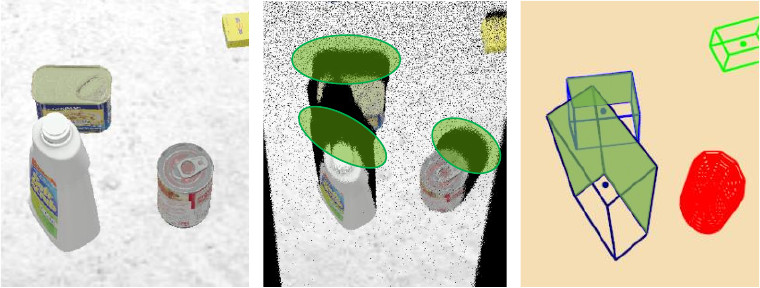}}
	\subfigure[Information gain under different camera views.]{\includegraphics[scale=0.45]{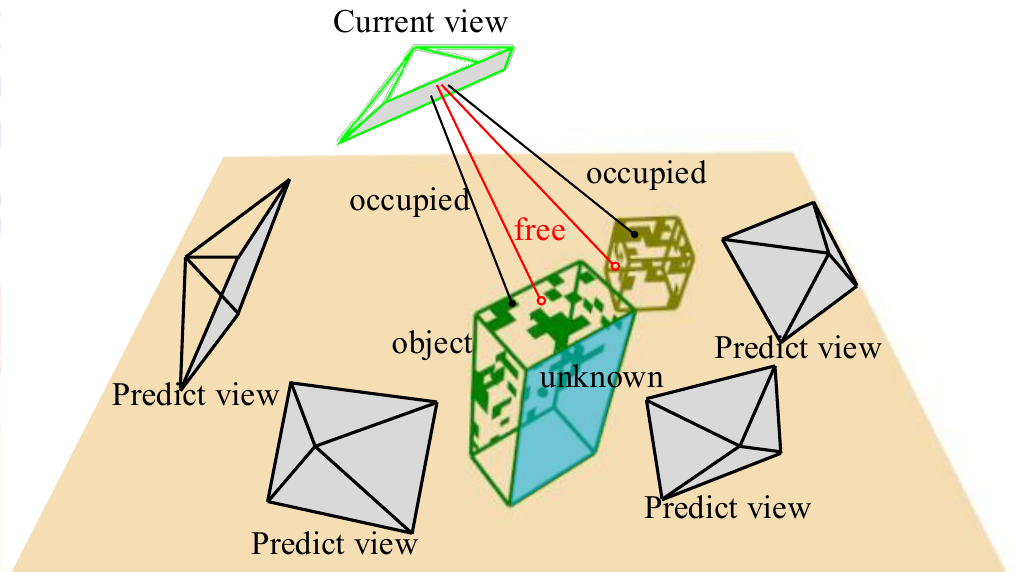}}
	\caption{Demonstration of the object-driven exploration.}
	\label{Perspective Prediction}
	\vspace{-4mm}
\end{figure}

\textbf{Information Gain Definition}: As illustrated in Fig. \ref{Perspective Prediction}(b), Object-driven exploration aims to predict the information gain of different candidate camera views and then select the one to explore that maximizes the information gain, i.e., the NBV. The information in this work is defined as the map's uncertainty as mentioned in Section \ref{OIM}. The information gain is thus defined as the measurement of uncertainty reduction and accuracy improvement after the camera is placed at a specific pose. Conventionally, information gain is defined based on the area of unknown regions in the environment, i.e., the black holes in the medium subfigure of Fig. \ref{Perspective Prediction}(a), which may mislead the object map building. Compared with the conventional one, our proposed information gain is built on the object's observation completeness measurement, shown in the right subfigure of Fig. \ref{Perspective Prediction}(a), and incorporates the influence on object pose estimation, which is one of the key contributions in this work. 

\textbf{Information Gain Modeling}: As indicated by the definition, information gain is contingent on many factors. Thus, we create a utility function to model the information gain by first manually designing a feature vector to parameterize the factors. The following is the design of the feature vector used to characterize the object $\mathbf{x}$,
\vspace*{-0.5\baselineskip}
\begin{equation}
\mathbf{x}=\left(H_{obj}, \bar{H}_{obj},  R_{o}, R_{IoU}, \bar{V}_{obj}, s\right),
\end{equation}  
where $H_{obj}$, and $\bar{H}_{obj}$ are defined by Eq. (\ref{eq:grid}) - (\ref{eq:ba_grid}), $R_o$ is the ratio of occupied grids to the total grids of the object, which indicates the richness of its surface texture, $R_{IoU}$ is the 2D mean IoU with adjacent objects used for modeling occlusion under a specific camera view,  $\bar{V}_{obj}$ is the current volume of the object, and $s$ is a binary value used for indicating whether the object is fully explored.

The utility function for NBV selection then is defined as,
\vspace*{-0.5\baselineskip}
\begin{equation}
\label{uti}
f = \sum_{\mathbf{x}\in I}  \left( (1-R_{o})H_{obj}+\lambda(H_{IoU}+H_{V})  \right) s(\mathbf{x}),
\end{equation}
where $I$ is the predict camera view, $\lambda$ is a weight coefficient ($\lambda=0.2$ in our implementation), and $H_{IoU}, H_{V}$ share the same formula,
\vspace*{-0.5\baselineskip}
\begin{equation}
\label{eq:plogp}
H=- p \log (p).
\end{equation}

The first item $\sum_{\mathbf{x}\in I}(1-R_{o})H_{obj}$ in Eq. \eqref{uti} is used to model the total weighted uncertainty of the object map under the predict camera view. Here we give more weight not only to the unknown grids but also the free ones by using $1-R_{o}$. The reason is that we want to encourage more explorations in free regions to find more image features that are neglected by previous sensing. 

The second item $\sum_{\mathbf{x}\in I}H_{IoU}$ in Eq. \eqref{uti} defines the uncertainty on object detection, which is one of the key factors affecting object pose estimation (see Fig. \ref{Pose Estimation}). The uncertainty is essentially caused by occlusions between objects. We use this item to encourage a full observation of the object. The variable in Eq. \eqref{eq:plogp} is the rescaled 2D IoU, i.e., $p=R_{IoU}/2$.

The third item $\sum_{\mathbf{x}\in I}H_{V}$ in Eq. \eqref{uti} models the uncertainty on object pose estimation. Under different camera views, the estimated object poses are usually different, and consequently, induce the changes in object volume. Here, we first fit a standard normal distribution using the normalized history volumes $\{\bar{V}_{obj}^{(0)}, \bar{V}_{obj}^{(1)}, \cdots, \bar{V}_{obj}^{(t)}\}$ of each object, and then take the probability density of $\bar{V}_{obj}^{(t)}$ as the $p$ value in Eq. \eqref{eq:plogp}. This item essentially encourages the camera view that can converge the pose estimation process.

The $s(\mathbf{x})$ in Eq. \eqref{uti} indicates whether the object should be considered during the calculation of the utility function. $s(\mathbf{x})\small{=}0$, if the following condition is satisfied: $(\bar{H}_{grid} < 0.5 \lor R_o > 0.5)  \land p(\bar{V}_{obj}^{(t)}) > 0.8$. If this condition holds for all the objects, or the maximum tries is achieved (10 in this work), the exploration will be finished.

Based on the utility function, the NBV that maximizes $f$ is continuously selected and leveraged to guide the exploration process, during which the global object map is also incrementally built up, as indicated in Fig. \ref{System_Framework}.

\section{Experimental Results}

\begin{table}[tb]\small
	\vspace{2mm}
	\centering
	\renewcommand{\arraystretch}{1.3}
	\setlength{\abovecaptionskip}{-0.02cm}
	\caption{ACCURACY OF OBJECT POSE ESTIMATION.}
	\begin{center}
		\label{table1}
		\begin{tabular}{c|c|cccc}
			\hline
			\multicolumn{1}{l|}{Scene} & Metrics   & Ours          & Random.       & Cover.           & Init.        \\ \hline
			
			& 3D IoU       & \textbf{0.427}  & 0.3056       & 0.3329          & 0.3586       \\
			
			& 2D IoU  & \textbf{0.6225} & 0.4571       & 0.5221          & 0.5212       \\
			
			& CDE & \textbf{1.5272} & 2.2699       & 1.7876          & 2.2022       \\
			
			\multirow{-4}{*}{1}        & YAE    & 3.5             & 4.8          & 3.8             & \textbf{2.8} \\ \hline
			
			& 3D IoU       & \textbf{0.4307} & 0.3017       & 0.4224          & 0.3400       \\
			
			& 2D   IoU    & \textbf{0.8679} & 0.6422       & 0.7730          & 0.6480       \\
			
			& CDE & \textbf{1.4646} & 2.1096       & 1.5931          & 1.9822       \\
			
			\multirow{-4}{*}{2}        & YAE & 2.4             & \textbf{1.8} & 2.7             & 2.4          \\ \hline
			
			& 3D IoU       & \textbf{0.4132} & 0.3125       & 0.3685          & 0.2617       \\
			
			& 2D   IoU    & \textbf{0.6225} & 0.4915       & 0.5517          & 0.3909       \\
			
			& CDE & 1.5503          & 2.0672       & \textbf{1.4841} & 2.7489       \\
			
			\multirow{-4}{*}{3}        & YAE  & 3.9             & \textbf{3.7} & 3.8             & 4.9          \\ \hline
			
			& 3D IoU       & \textbf{0.4790} & 0.3824       & 0.3664          & 0.3007       \\
			
			& 2D IoU  & \textbf{0.6536} & 0.5886       & 0.4788          & 0.4869       \\
			
			& CDE & \textbf{1.3335} & 1.3514       & 1.7508          & 1.927        \\
			
			\multirow{-4}{*}{4}        & YAE       & 2.9             & 2.8          & \textbf{2.1}    & \textbf{2.1} \\ \hline
			
			& 3D IoU       & \textbf{0.5177} & 0.2696       & 0.2884          & 0.3720       \\
			
			& 2D IoU      & \textbf{0.6263} & 0.4297       & 0.4326          & 0.6142       \\
			
			& CDE & \textbf{1.3704} & 2.5077       & 2.1753          & 2.0084       \\
			
			\multirow{-4}{*}{5}        & YAE  & 3.9             & \textbf{2.1} & 3.9             & \textbf{2.1}          \\ \hline
			
			& 3D IoU       & \textbf{0.4411} & 0.3000       & 0.3597          & 0.2916       \\
			
			& 2D IoU    & \textbf{0.5437} & 0.4850       & 0.4783          & 0.5042       \\
			
			& CDE & \textbf{2.5998} & 3.4278       & 2.8411          & 3.4965       \\
			
			\multirow{-4}{*}{6}        & YAE    & \textbf{2.3}    & 2.7          & 3               & 2.7          \\ \hline
			
			& 3D IoU       & \textbf{0.4626} & 0.2118       & 0.4133          & 0.3153       \\
			
			& 2D IoU    & \textbf{0.6017} & 0.3839       & 0.5569          & 0.4541       \\
			
			& CDE & 1.49928         & 2.3822       & \textbf{1.4832} & 2.0467       \\
			
			\multirow{-4}{*}{7}        & YAE    & \textbf{2.1}    & 4.5          & 2.5             & 2.5          \\ \hline
			
			& 3D IoU       & \textbf{0.453}  & 0.2977       & 0.3645          & 0.3200       \\
			
			& 2D IoU      & \textbf{0.6483} & 0.4969       & 0.5419          & 0.5171       \\ 
			
			& CDE & \textbf{1.6207} & 2.3022       & 1.8736          & 2.3446       \\
			
			\multirow{-4}{*}{Mean}     & YAE   & 3               & 3.2          & 3.1             & \textbf{2.8} \\ \hline
		\end{tabular}
	\end{center}
	\vspace{-10mm}
\end{table}

To demonstrate the effectiveness of the map and the viability of robot manipulation guided by the map. we conduct extensive evaluations in both simulation and real-world settings. In which, the simulated robotic manipulation scenes are set in Sabine \cite{xiang2020sapien}, shown in Fig. \ref{Mapping_Result}, the number of objects and the scene complexities vary in different scenes. The physical experiments are carried out on the Kinova MICO robotic arm. All experiments are conducted in real-time on a PC equipped with a GTX1070 GPU.

\begin{figure}[htbp]
	\vspace{2mm}
	\centering
	%	\captionsetup{belowskip=-10pt}
	\includegraphics[scale=0.36]{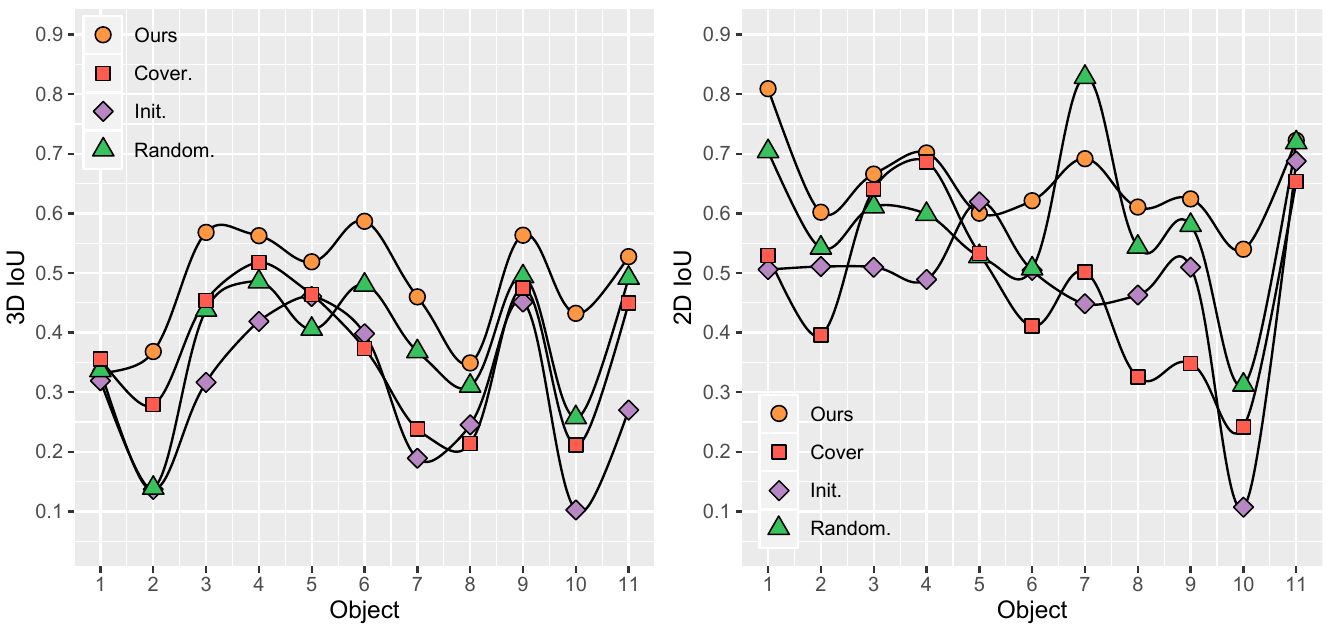}
	\caption{3D IoU and 2D IoU estimated by different strategies.}
	\label{IoU}
	\vspace{-8mm}
\end{figure}

\subsection{Evaluation of Object Pose Estimation}

\begin{figure*}[t]
	\centering
%	\captionsetup{belowskip=-10pt}
	\includegraphics[scale=0.31]{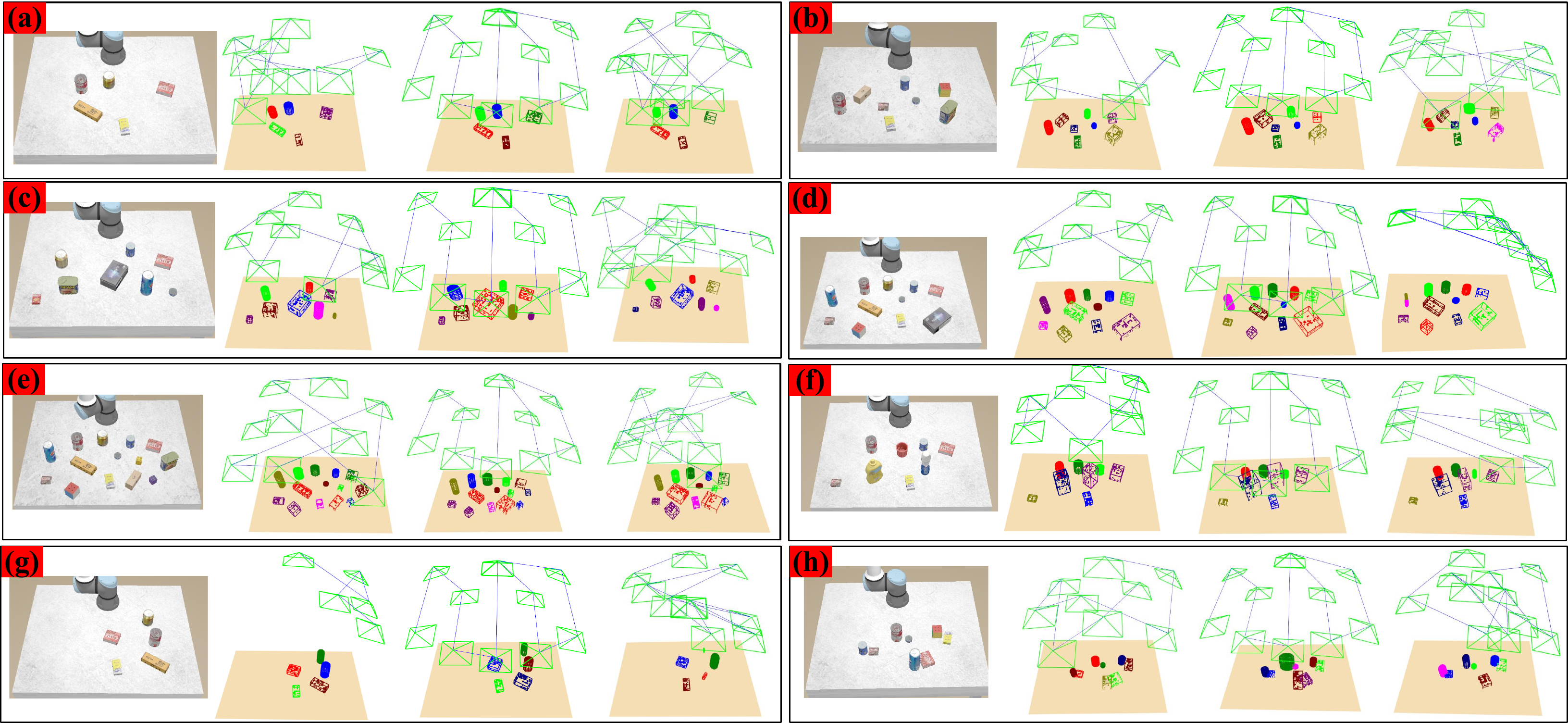}
	\caption{Comparison of mapping results. The first column in the sub-picture: the scene image; the second column: the result of our object-driven exploration; the third column: the result of the coverage exploration; the fourth column: the result of the randomized exploration.}
	\label{Mapping_Result}
	\vspace{-4mm}
\end{figure*}

The accurate position estimate is critical for successful robotic manipulation operations such as grasping, placing, arranging, and planning. However, precision is difficult to ensure when the robot estimates autonomously. To quantify the effect of active exploration on object pose estimation, like previous studies \cite{arruda2016active,morrison2019multi}, we compare our object-driven method with two typically used baseline strategies, i.e., randomized exploration and coverage exploration. As indicated in Fig. \ref{Mapping_Result}, for randomized exploration, the camera pose is randomly sampled from the reachable set relative to the manipulator, while for coverage exploration, a coverage trajectory based on Boustrophedon decomposition \cite{kaljaca2020coverage} is leveraged to scan the scene. At the beginning of all the explorations, an initialization step, in which the camera is sequentially placed over the four desk corners from a top view, is applied to start the object mapping process. The simulator provides the ground truth of object position, orientation, and size. Correspondingly, the accuracy of pose estimation is evaluated by the Center Distance Error (CDE, c$\rm m$), the Yaw Angle Error (YAE, $\rm degree$), and the IoU (including 2D IoU from the top view and 3D IoU) between the ground truth and our estimated results.

Table \ref{table1} shows the evaluation results in seven scenes (Fig. \ref{Mapping_Result}). We can see our proposed object-driven exploration strategy achieves a \textbf{3D IoU} of 45.3\%, which is 15.53\%, 8.85\%, and 13.3\% higher than that of the randomized exploration, the coverage exploration, and the initialization, respectively. For \textbf{2D IoU}, our method achieves an accuracy of 64.83\%, which is 15.14\%, 10.64\%, and 13.12\% higher than the baseline methods and initialization, respectively. In terms of \textbf{CDE}, our method reaches 1.62cm, significantly less than other methods. For \textbf{YAE}, all exploration strategies achieve an error of approximately ${3}^{\circ}$, which verifies the robustness of our line-alignment \cite{wu2020eao} and the yaw angle optimization method (Eq. (\ref{yaw1})). Additionally, the \textbf{pose refinement method} described in section \ref{section: Observation} and Fig. \ref{Pose Refine} achieves a reduction of roughly 0.2cm in object size inaccuracy, which results in a reduction of around 0.1cm in CDE. The level of precision attained is sufficient for robotic manipulation\cite{wang2019densefusion}. Moreover, we also find that randomized exploration sometimes performs worse than the initialization result (rows 2, 5, and 7), indicating that increasing observations not necessarily results in more accurate pose estimation, and purposeful exploration is necessary.%\hl{targeted exploration is necessary.}

Fig. \ref{IoU} shows the position estimation results of 11 objects in Fig. \ref{Mapping_Result}(d), in which the 2D IoU from the top view and the 3D IoU are reported. It can be seen that our proposed strategy achieves higher 3D IoU and 2D IoU scores compared with the baseline methods. 

\begin{figure*}[htbp]
	\centering
	\subfigure[Grasping process in the simulated environment.]{\includegraphics[scale=0.34]{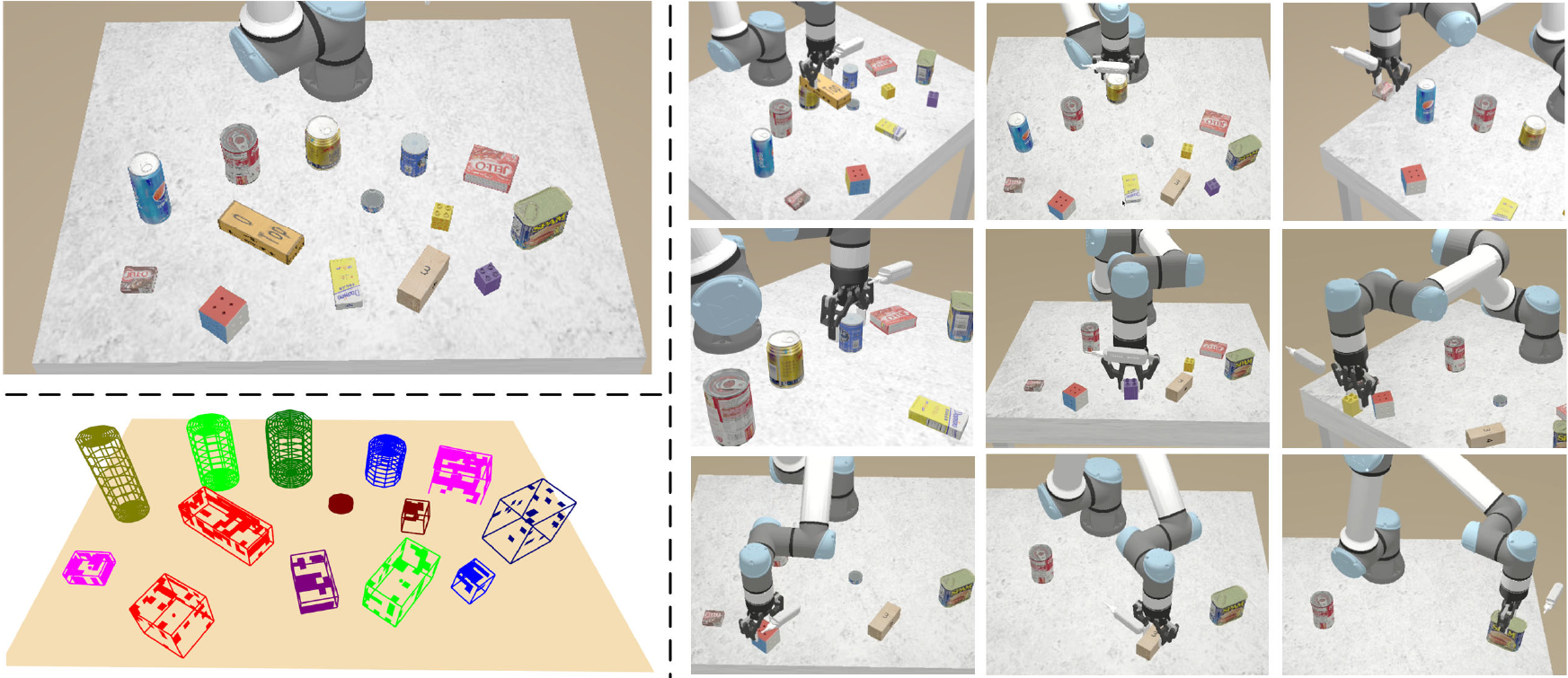}}
	\subfigure[Grasping process in the real world.]{\includegraphics[scale=0.23]{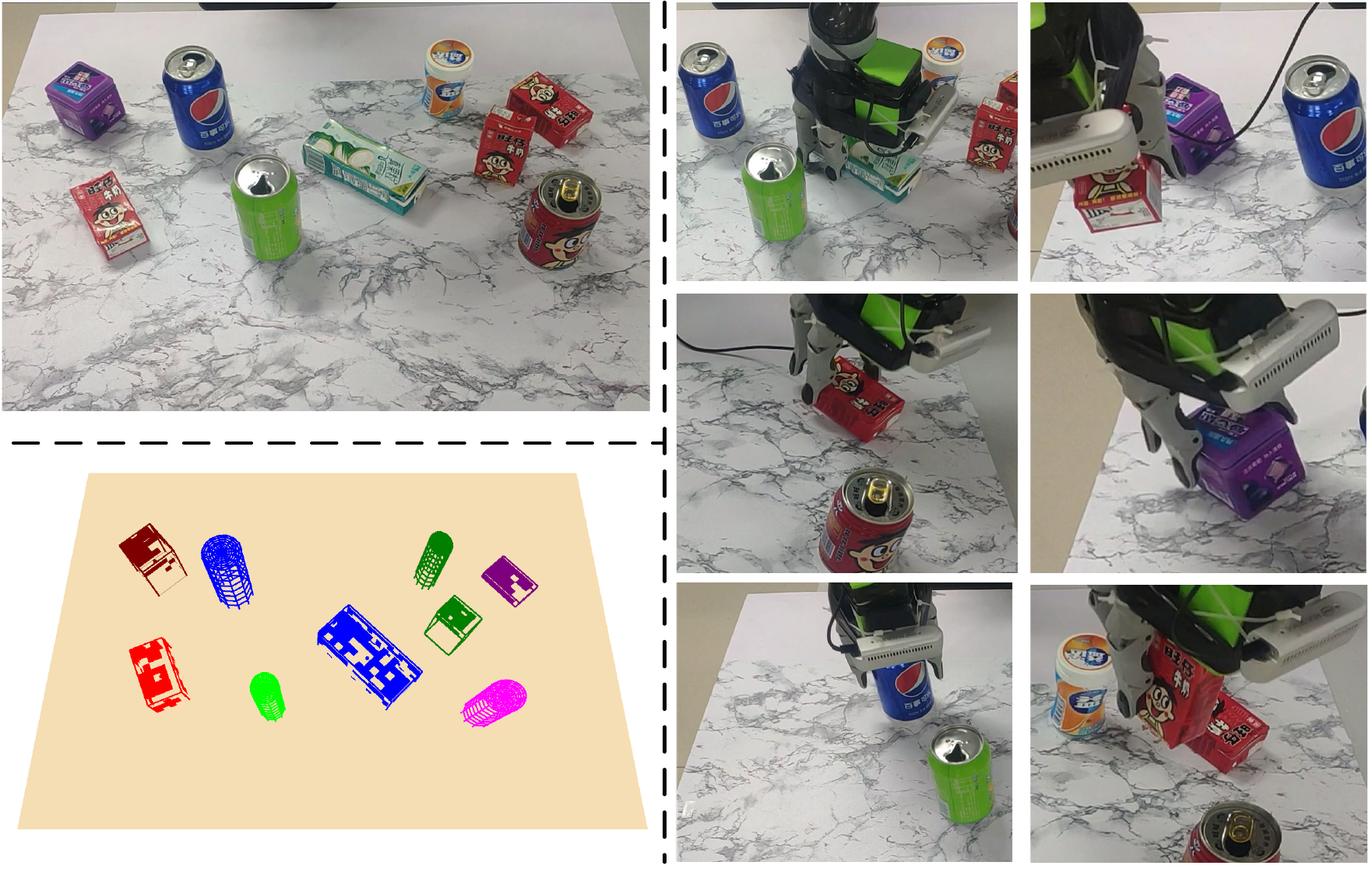}}
	\caption{The demonstration of grasping. Left-top (in each sub-figure): grasping scene. Left-bottom: object map. Right: part of the grasping process.}
	\label{Grasp_Experiments}
	\vspace{-5mm}
\end{figure*}

\subsection{Evaluation of Object Map Quality}
\vspace{-2mm}

The mapping results are demonstrated in Fig. \ref{Mapping_Result}. The cubes and cylinders are used to model the objects, including poses and scales (analyzed above), based on their semantic categories. The following characteristics are present: \textbf{1)} The system is capable of accurately modeling a variety of objects as the number of objects increases, shown in Fig. \ref{Mapping_Result}(a)-(e), which demonstrates its robustness.
\textbf{2)} Among objects with various sizes, our method focuses more on large objects with lower observation completeness (see Fig. \ref{Mapping_Result}(f)). 
\textbf{3)} When objects are distributed unevenly, our proposed strategy can swiftly concentrate the camera on object regions, thus avoiding unnecessary and time-consuming exploration (see Fig. \ref{Mapping_Result}(g).
\textbf{4)} For scenes with objects close to each other, our method can place more attention on regions with fewer occlusions (see Fig. \ref{Mapping_Result}(h)). These behaviors verify the effectiveness of our exploration strategy. Additionally, our method has a shorter exploration path yet produces a more precise object posture, demonstrating its efficiency.

\subsection{Object Grasping Experiment}
\label{section: Object Grasping}

In this experiment, we leverage the incrementally generated object map to perform object grasping. Fig. \ref{Grasp_Experiments}(a) and Fig. \ref{Grasp_Experiments}(b) illustrate the grasping process in simulated and real-world environments, respectively, with the object map included. After extensive testing, we obtained a grasping success rate of approximately 86\% in the simulator and 81\% in the real world, which may be affected by environmental or manipulator noises.

It is found that the center and direction of the objects have a significant influence on grasping performance. Our proposed method performs well in terms of these two metrics, thus ensuring high-quality grasping. According to our experiments, a miscalculation of the object's height is another cause of gripping failure. This is because the robot tends to observe objects from the top view, which is prone to depth inaccuracies. This is also the reason why 2D IoU is generally higher than 3D IoU, implying the necessity of multi-angle exploration. Overall, our pose estimation results can meet the requirements of object grasping.

\begin{figure}[t]
	\centering
%	\captionsetup{belowskip=-10pt}
	\includegraphics[scale=0.3]{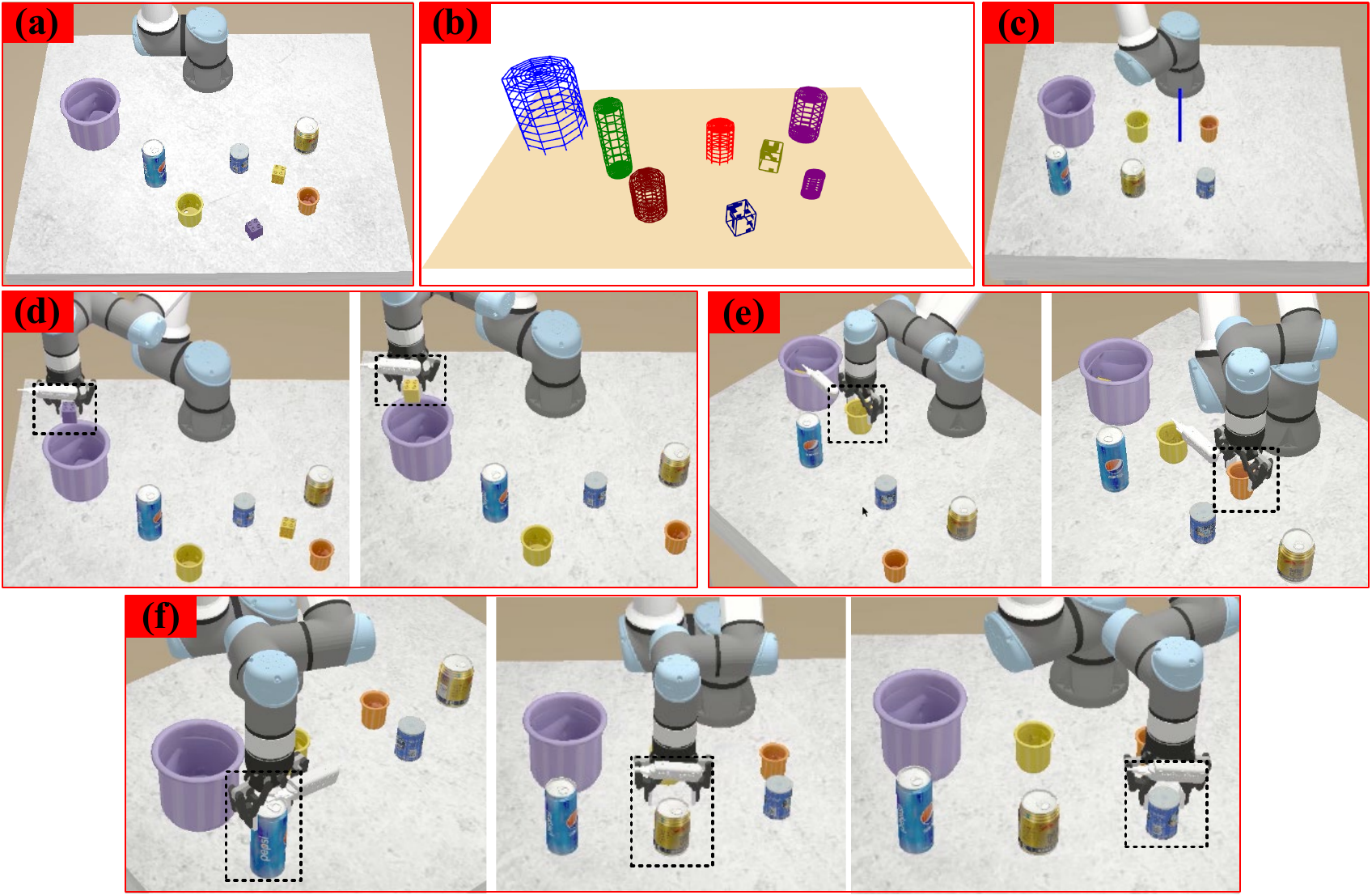}
	\caption{Object placement according to the global object map. (a): initial scene.  (b): object map. (c): target scene. (d)-(f): object placement process.}
	\label{Placement}
	\vspace{-6mm}
\end{figure}

\subsection{Object Placement Experiment}
\label{section: Object Placement}

We argued that the proposed object map level perception outperforms object pose-only perception, providing information for more intelligent robotics decision-making tasks in addition to grasping. Such include avoiding collisions with other objects, updating the map after grasping, object arrangement and placement based on object properties, and object delivery requested by the user. To verify the global perception capabilities introduced by object mapping, we design the object placement experiments.
As shown in Fig. \ref{Placement}, The robot is required to perform manipulation from the original scene (see Fig. \ref{Placement}(a)) to the target scene (see Fig. \ref{Placement}(c)) according to object sizes and classes encoded in the object map.

The global object map is shown in Fig. \ref{Placement}(b), which contains the semantic labels, size, and pose of the objects. The two little blocks are picked up and placed in the large cup (Fig. \ref{Placement}(d)); the cups are sorted by volume (Fig. \ref{Placement}(e)), and the bottles are arranged by height (Fig. \ref{Placement}(f)). This is challenging for the conventional grasping approach since lacking global perception such as the level of the object on the map, its surroundings, and could interact with which objects.

\begin{figure}[htbp]
	\centering
	\includegraphics[scale=0.46]{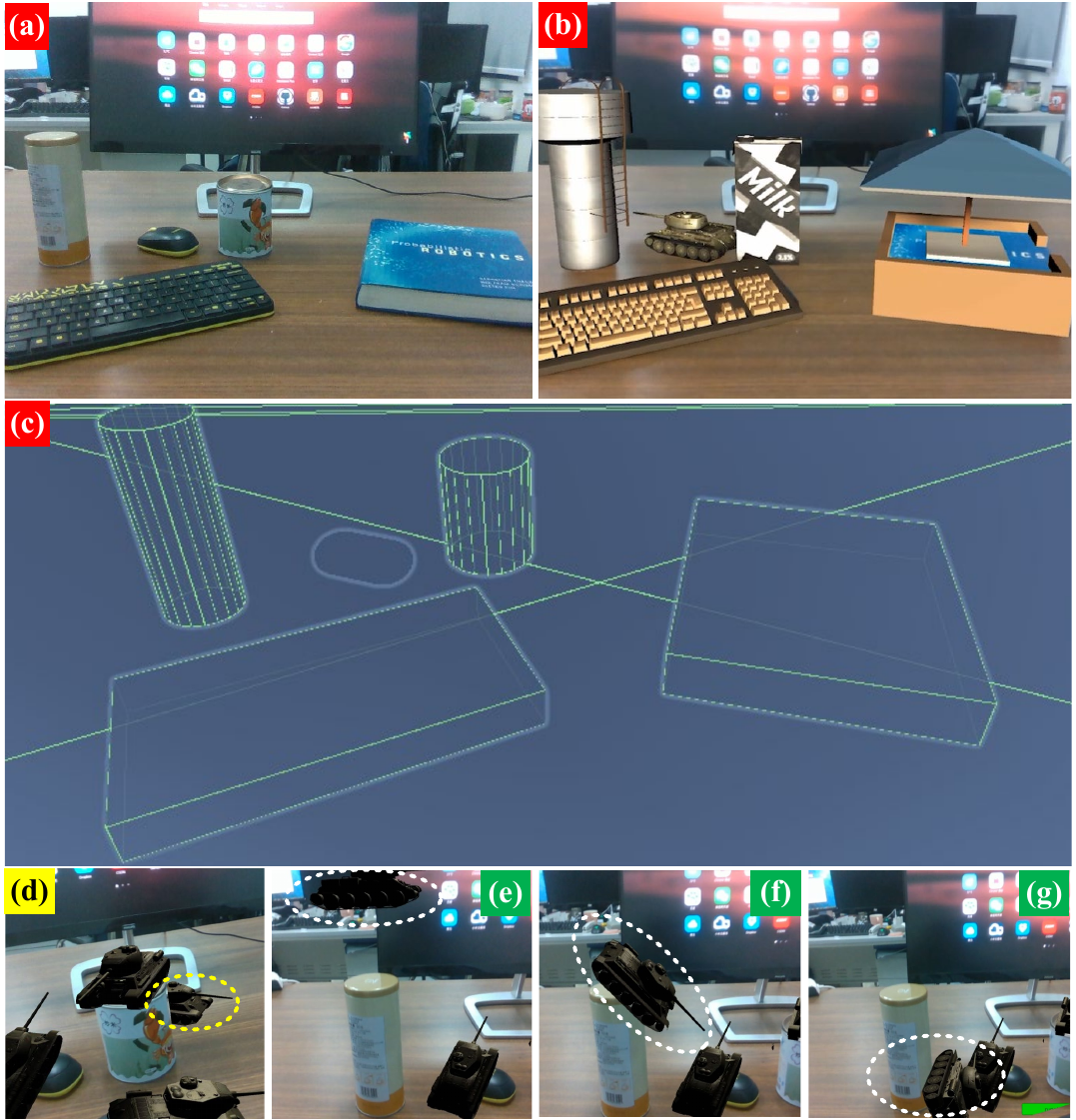}
	\caption{Augmented Reality experiments. (a): original scene. (b): augmented reality scene. (c) object map. (d):  occlusion effect. (e)-(g): collision effect.}
	\label{Augmented_Reality}
	\vspace{-6mm}
\end{figure}

\subsection{Augmented Reality Experiment}

This is an extended experiment beyond robot manipulation. Conventional augmented reality focus on localization and 3D register. The proposed SLAM-based object map promotes augmented reality to the next level, in which the object map provides more complete environment information for augmented reality, allowing for a more realistic immersive experience \cite{zhou2016pmomo,holynski2018fast,sun2021neuralrecon}. Here, we add virtual objects in Fig. \ref{Augmented_Reality}(a) to extend the proposed method in this work and further demonstrate the effectiveness of pose estimation.

From the experimental results, the desktop scene is first modeled, and an object-level map is then obtained (see Fig. \ref{Augmented_Reality}(c)). Fig. \ref{Augmented_Reality}(b) illustrates 3D object registration, in which virtual models are positioned in the correct position based on the estimated object pose and size.  Fig. \ref{Augmented_Reality}(d) demonstrates the occlusion effect, in which the virtual tank behind the cup is partially observed, and the tank above the cup can still be fixed in space. Fig. \ref{Augmented_Reality}(e)-(g) show the collision effect, in which the tank falls and collides with the cup, and finally drops on the desktop. Which is challenging for the traditional method since the low perception ability of the environment.

\section{Conclusion}
\vspace{-1mm}
In this paper, we for the first time present an object-driven active mapping framework based on an object SLAM with the suggested observation completeness measurement method and the object-driven active exploration strategy. The framework aims to actively enable the full observation and accurate pose estimation of unknown objects, and then implement complex robotic manipulation and autonomous perception tasks. Extensive experiments significantly verify the effectiveness of our proposed framework. The presented idea and methods in this work will significantly push forward related studies in this research area. The major limitation of this work lies in that the framework currently cannot model complex objects with irregular shapes and clutter settings, which is also challenging for the object SLAM.

\section*{Acknowledgements}
\vspace{-1mm}
This work was supported by National Natural Science Foundation of China (No. 61973066, 61471110) , Fundation of Key Laboratory of  Aerospace System Simulation(6142002200301), Fundation of Key Laboratory of Equipment Reliability(61420030302), Fundamental Research Funds for the Central Universities(N182608004, N2004022) and Distinguished Creative Talent Program of Liaoning Colleges and Universities (LR2019027).

{\small
\bibliographystyle{ieee_fullname}
\bibliography{egbib}
}

\end{document}